\pdfoutput=1
\documentclass[11pt]{article}
\usepackage{authblk} % for fancy author format

\usepackage[utf8]{inputenc}
\usepackage{ACL2023}
\usepackage{hyperref}

\usepackage{times}
\usepackage{latexsym}

\usepackage{pgf}
\usepackage{import}
\usepackage{layouts}

\usepackage{multirow}
\usepackage{booktabs}

\usepackage{graphics}

\usepackage[T1]{fontenc}
\usepackage{microtype}
\usepackage[compact]{titlesec}

\title{The Ecological Fallacy in Annotation:\\Modelling Human Label Variation goes beyond Sociodemographics}

\author[1]{\textbf{Matthias Orlikowski}}
\author[2]{\textbf{Paul R\"ottger}}
\author[1]{\textbf{Philipp Cimiano}}
\author[3]{\textbf{Dirk Hovy}}
\affil[1]{Bielefeld University}
\affil[2]{University of Oxford}
\affil[3]{Computing Sciences Department, Bocconi University, Milan, Italy}

\begin{document}
\maketitle

\begin{abstract}
  Many NLP tasks exhibit human label variation, where different annotators give different labels to the same texts.
  This variation is known to depend, at least in part, on the sociodemographics of annotators.
  Recent research aims to model individual annotator behaviour rather than predicting aggregated labels, and we would expect that sociodemographic information is useful for these models.
  On the other hand, the ecological fallacy states that aggregate group behaviour, such as the behaviour of the \textit{average} female annotator, does not necessarily explain individual behaviour.
  To account for sociodemographics in models of individual annotator behaviour, we introduce group-specific layers to multi-annotator models.
  In a series of experiments for toxic content detection, we find that explicitly accounting for sociodemographic attributes in this way does not significantly improve model performance.
  This result shows that individual annotation behaviour depends on much more than just sociodemographics.
\end{abstract}

%%%%%%%%%%%%%%%%%%%%%%%%%%%%%%%%%%%%%%%%%%%%%%%%%%%%%%%%%%%%%%%%%%%%%%%%%%%%%%%%%%%%%%%%%%%
%%%%%%%%%%%%%%%%%%%%%%%%%%%%%%%%%%%%%%%%%%%%%%%%%%%%%%%%%%%%%%%%%%%%%%%%%%%%%%%%%%%%%%%%%%%

\section{Introduction}

Different annotators will not necessarily assign the same labels to the same texts,
resulting in human label variation \citep{plank-2022-problem}.
Previous work finds that this variation depends at least in part on the sociodemographics of annotators, such as their age and gender \citep{binns_like_2017,al-kuwatly-etal-2020-identifying,excell-al-moubayed-2021-towards,shen-rose-2021-sounds}.
These results are particularly pronounced for subjective tasks like toxic content detection 
\citep{sap-etal-2019-risk,kumar_designing_2021,sap-etal-2022-annotators, goyal_is_2022}.
Since human label variation is relevant to a wide range of NLP tasks, recent research has begun to model individual annotator behaviour, rather than predicting aggregated labels \citep{davani-etal-2022-dealing,gordon_jury_2022}.
In this setting, we would expect sociodemographic attributes to help explain annotator decisions. Therefore, we investigate \textbf{whether explicitly accounting for the sociodemographic attributes of annotators leads to better predictions of their annotation behaviour}\footnote{Code to run our experiments and analyses is available at \url{https://github.com/morlikowski/ecological-fallacy}}.

\begin{figure}
\centering
\includegraphics[width=75mm]{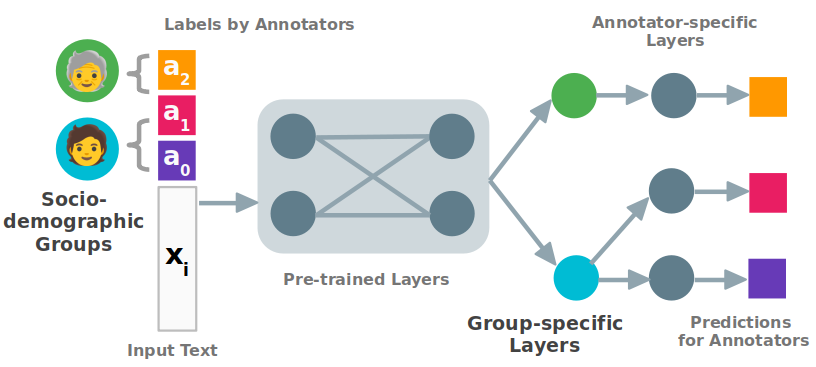}
\caption{Group-specific layers representing annotator sociodemographics in multi-annotator models.}
\label{fig:architecture}
\end{figure}

There is a risk of misreading these efforts as an example of the \textit{ecological fallacy}: aggregate group behaviour does not necessarily explain individual behaviour \citep{robinson_ecological_1950, freedman_ecological_2015}.
For example, while on average, white annotators may be more likely to label African-American Vernacular English as toxic \citep{sap-etal-2019-risk}, that does not mean it is true for \textit{every} white annotator individually. However, we aim at exactly this distinction to discuss the relevance of sociodemographic groups in models of individual annotator behaviour. Likewise, we do not assume prior work to commit ecological fallacies, even if a less-nuanced read might suggest it.

\citet{davani-etal-2022-dealing} introduce a simple multi-annotator model, where each annotator is modelled with a separate classification head.
We expand their model with \textit{group-specific} layers, which are activated for each annotator based on their sociodemographic attributes.
We compare the two model setups to a control setup where we randomise group assignments. All comparisons use annotator-level toxicity data from \citet{kumar_designing_2021}.
We find that find that explicitly accounting for sociodemographic attributes does \textit{not} significantly improve model performance. This result suggests that human label variation happens at a more individual level than sociodemographics, and that annotator decisions are even more complex.

\paragraph{Contributions}

1)~We introduce group-specific layers to model groups of annotators with shared attributes in multi-annotator models.
2)~We evaluate the effect of group-specific layers for toxic content detection, and show that explicitly accounting for sociodemographic attributes does not significantly improve performance, thus highlighting the risk of the ecological fallacy in annotator modelling.

As a corollary, we show that multi-annotator models can be applied to many times more annotators than in prior work.

%%%%%%%%%%%%%%%%%%%%%%%%%%%%%%%%%%%%%%%%%%%%%%%%%%%%%%%%%%%%%%%%%%%%%%%%%%%%%%%%%%%%%%%%%%%
%%%%%%%%%%%%%%%%%%%%%%%%%%%%%%%%%%%%%%%%%%%%%%%%%%%%%%%%%%%%%%%%%%%%%%%%%%%%%%%%%%%%%%%%%%%

\section{Related Work}
 
\paragraph{Sociodemographics in Annotation Behaviour}
\label{sociodemographics_in_annotation}

A growing body of research studies how annotator sociodemographics relate to their annotation decisions, for tasks ranging from natural language inference \citep{biester-etal-2022-analyzing} to the detection of racist \citep{larimore-etal-2021-reconsidering} or generally toxic \citep{sap-etal-2022-annotators} language.
\citet{goyal_is_2022}, for example, find that annotators from certain sociodemographic groups (e.g., LGBTQ people) tend to find content attacking their own groups (e.g., homophobic content) to be more toxic.
This motivates our research into explicitly accounting for sociodemographics to model annotation behaviour.
However, the link between sociodemographics and behaviour is not uncontested.
\citet{biester-etal-2022-analyzing}, for example, do not find significant differences in annotation behaviour between annotators of different genders for four different tasks.

\paragraph{Predicting Annotators' Decisions on Text}
\label{sec:predicting_individual_annotators}
Different from analyses of annotation behaviour, a recent line of research attempts to learn models based on individual annotations \citep{plank-etal-2014-learning,jamison-gurevych-2015-noise,akhtar_modeling_2020,fornaciari-etal-2021-beyond,cercas-curry-etal-2021-convabuse}. These models are motivated by the concern that aggregating labels into a single ``truth'' is too simplistic for many tasks \citep{uma_learning_survey_2021,basile-etal-2021-need} and might introduce uneven representation of perspectives \citep{prabhakaran-etal-2021-releasing, nlperspectives-2022-perspectivist}.

A particular way of learning from disaggregated labels are models that predict individual annotator decisions for an example. Our work builds directly on such a model, multi-annotator models \citep{davani-etal-2022-dealing}, which we describe in more detail separately (\S\ref{sec:experiments}). \citet{gordon_jury_2022} present a model which also predicts individual annotations and allows a user to interactively aggregate them based on ``a jury'' inspired by the US judicial system. Their work is similar to ours in central aspects as they explicitly model annotators' sociodemographics and use the same dataset as we do \citep{kumar_designing_2021}. 
Different from our work, they frame the task as 
a regression problem and develop a model based on recommender systems. While they also explore ecological fallacies, they focus on usage risks of their system and countermeasures. In contrast, we consider the issue of the ecological fallacy in modelling annotation behaviour more generally. We compare our findings to their results (\S\ref{sec:discussion}).

%%%%%%%%%%%%%%%%%%%%%%%%%%%%%%%%%%%%%%%%%%%%%%%%%%%%%%%%%%%%%%%%%%%%%%%%%%%%%%%%%%%%%%%%%%%
%%%%%%%%%%%%%%%%%%%%%%%%%%%%%%%%%%%%%%%%%%%%%%%%%%%%%%%%%%%%%%%%%%%%%%%%%%%%%%%%%%%%%%%%%%%

%%%%%%%%%%%%%%%%%%%%%%%%%%%%%%%%%%%%%%%%%%%%%%%%%%%%%%%%%%%%%%%%%%%%%%%%
\section{Data} \label{sec:data}

We use a sample of the \citet{kumar_designing_2021} dataset for our experiments. The full dataset contains 107,620 English comments from Twitter, Reddit, and 4Chan, annotated for toxicity by 17,280 annotators. The annotation process encouraged annotator subjectivity \citep{roettger-etal-2022-two} which is a desired feature for modelling annotator behaviour.
For each annotator, there is extensive sociodemographic information, collected with a survey. Annotations are given as ratings on a five-point scale which we convert to binary annotations by mapping ratings of 2 to 4 to \textit{toxic}, and ratings 0 and 1 to \textit{non-toxic}.

We randomly sample comments from the dataset until we reach annotations from more than 5,000 annotators.
We then add all other annotations by these annotators.
This approach maximizes the number of examples while controlling the number of annotators in our sample.

Our final sample contains 111,780 annotations from 5,002 annotators on 22,360 comments with 20 to 120 annotations per annotator (mean 22.35). Most comments have five annotations. 20 comments have four because we removed any underage annotators before sampling. In total 78,357 annotations (70.10\%) are toxic, and 33,423 annotations (29.90\%) are non-toxic.

We focus on four sociodemographic attributes: gender, age, education, and sexual orientation. Group sizes vary by attribute. For gender, 2,450 annotators (48.98\%) identify as female, 2,116 (42.30\%) as male, 23 (0.46\%) as non-binary (rest in residual categories, full statistics in \ref{sample-sociodemographics}).

%%%%%%%%%%%%%%%%%%%%%%%%%%%%%%%%%%%%%%%%%%%%%%%%%%%%%%%%%%%%%%%%%%%%%%%%%%%%%%%%%%%%%%%%%%%
%%%%%%%%%%%%%%%%%%%%%%%%%%%%%%%%%%%%%%%%%%%%%%%%%%%%%%%%%%%%%%%%%%%%%%%%%%%%%%%%%%%%%%%%%%%
\section{Experiments}
\label{sec:experiments}
We compare three models. The \textbf{baseline} model is the multi-annotator model by \citet{davani-etal-2022-dealing}. We use their multi-task variant:
For each annotator, there is a separate classification layer trained on annotations from that annotator. All annotator layers share a pre-trained language model used to encode the input.
We use RoBERTa \citep{liu2019roberta}
for this, motivated by computational constraints. 
The other models in our experiments build on this baseline model.

For the \textbf{sociodemographic} models, we add group-specific layers based on sociodemographic attributes of the annotators. A single attribute, e.g., age, implies several groups, e.g., \textit{ages 25-34}, \textit{ages 35-44}.
We add the group-specific layers between the pre-trained model and the annotator layers. Each group of annotators shares a separate group-specific layer.
We implement group-specific layers as fully-connected, linear layers, each learning a feature transformation applied for one group of annotators.

Finally, for the \textbf{random} models, we shuffle the assignment of annotators to groups from the sociodemographic model, retaining the relative group sizes.
In other words, the probability of each annotator staying in the same group or being reassigned to another group corresponds to the relative size of each group.
This approach keeps the model architecture constant while removing the connection between actual sociodemographic attributes and group assignment. It allows us to distinguish the effects of additional parameters, which group-specific layers add in comparison to the baseline, from the effects of sociodemographic information.

%%%%%%%%%%%%%%%%%%%%%%%%%%%%%%%%%%%%%%%%%%%%%%%%%%%%%%%%%%%%%%%%%%%%%%%%
\subsection{Evaluation Setup} \label{subsec: evaluation setup}

We evaluate all models on individual annotations from gender, age, education, and sexual orientation groups. This setup is comparable to the ``individual label'' evaluations in \citet{davani-etal-2022-dealing} and \citet{gordon_jury_2022}, but with scores calculated per group of annotators. We measure performance in macro-average $F_{1}$, to weigh each class equally.

\paragraph{Cross-Validation}
As there is no standard split available for our dataset, we perform three iterations of a four-fold cross-validation with different seeds (training details in Appendix \ref{appendix:training_details}). We choose four folds, so that even very small groups have more than a hundred annotations in each test set. Across folds, the numbers of annotations per sociodemographic group are similar (see Appendix \ref{appendix:support}).
We construct test sets that only contain comments unseen by the annotators in the training set. We also ensure that all test sets have similar proportions of toxic or non-toxic comments (assigned by the majority of annotators) to address the class imbalance in the dataset (70.62\% toxic, see \S\ref{sec:data}).

\paragraph{Statistical Significance}
We test for statistical significance  of our results from multiple runs of k-fold cross-validation via replicability analysis \citep{dror-etal-2017-replicability}.
We report the number of significant folds and the Bonferroni-corrected count \citep{dror-etal-2018-hitchhikers} in Appendix \ref{appendix:significance_tests}.
We compute the p-values for each fold via a paired bootstrap-sampling test with BooStSa \citep{fornaciari-etal-2022-hard}.
We set the significance level $\alpha = 0.05$, draw 1000 bootstrap samples per fold, and use a sample size of 50\% of the respective test set.

\paragraph{Remarks on Groups}
Annotators from different groups of the same attribute will in most cases not have annotated the same examples. Therefore, comparisons between models are only meaningful within each group.

The groups modeled via group-specific layers and those in the result tables are always the same. For example, if we report scores for gender groups, then the sociodemographic and randomized models are also based on gender groups.
In the following, we focus on a subset of groups, omitting, e.g., "Prefer not to say" (see Appendix \ref{app:full_results}).

%%%%%%%%%%%%%%%%%%%%%%%%%%%%%%%%%%%%%%%%%%%%%%%%%%%%%%%%%%%%%%%%%%%%%%%%
\section{Results} \label{subsec: results}
Table \ref{tab:results} shows the results for gender, age, education, and sexual orientation. A naive majority class baseline that predicts all input to be toxic performs worse than all other models with a large margin (exact results in Appendix \ref{app:full_results}).

\paragraph{Sociodemographics vs.\ Baseline}
Across attributes, the average scores of the sociodemographic model and the baseline are similar. 
The sociodemographic model often has a slightly higher average macro F1 than the baseline, but no statistically significant gains.
Where average performance is better by several points, as for homosexual annotators, this gain is offset by a large variance in performance (a consequence of small group sizes).

\paragraph{Sociodemographics vs.\ Random}
We also do not find significant performance differences between sociodemographic group-layer models and the corresponding random group assignment models.
For most groups, the randomized models achieve the highest average scores, but differences to the sociodemographic model are never statistically significant.

\setlength{\tabcolsep}{2pt}
\begin{table}[ht]
  \centering
  \small
  \begin{tabular}{llll}
\toprule
{Gender} & {Baseline} & {Soc-Dem.} & {Random} \\
\midrule
Male & \bfseries 68.00±0.49 & 67.66±0.46 & 67.63±0.53 \\
Female & 62.23±0.53 & 62.25±1.19 & \bfseries 62.41±0.92 \\
Nonbinary & 56.33±6.00 & 56.80±7.24 & \bfseries 58.00±7.49 \\
\bottomrule
\end{tabular}

  \vspace*{1em}

  \begin{tabular}{llll}
\toprule
{Age} & {Baseline} & {Soc-Dem.} & {Random} \\
\midrule
18 - 24 & 59.39±1.58 & 60.44±1.05 & \bfseries 60.52±1.37 \\
25 - 34 & 66.72±0.56 & 66.63±0.83 & \bfseries 66.92±0.51 \\
35 - 44 & 64.50±0.59 & 64.94±1.33 & \bfseries 65.24±0.89 \\
45 - 54 & 65.68±0.66 & 65.88±1.39 & \bfseries 65.98±0.83 \\
55 - 64 & 64.37±1.22 & \bfseries 64.94±1.66 & 64.84±1.30 \\
65 or older & 63.34±2.07 & \bfseries 64.70±2.21 & 62.77±2.39 \\
\bottomrule
\end{tabular}

   \vspace*{1em}

  \begin{tabular}{llll}
\toprule
{Education} & {Baseline} & {Soc-Dem.} & {Random} \\
\midrule
Associate degree & 60.69±1.44 & 60.54±2.35 & \bfseries 60.78±1.62 \\
Bachelor's degree & 66.16±0.51 & 66.23±0.82 & \bfseries 66.80±0.54 \\
Doctoral degree & 61.93±3.82 & \bfseries 63.79±5.03 & 63.27±3.67 \\
High school & 60.53±1.39 & 60.47±2.22 & \bfseries 60.55±1.87 \\
Below high school & 58.28±4.68 & \bfseries 62.12±4.90 & 60.17±4.25 \\
Master's degree & \bfseries 69.71±0.86 & 69.58±0.93 & 69.45±0.96 \\
Professional degree & 66.75±2.37 & 67.84±3.32 & \bfseries 68.62±2.84 \\
College, no degree & 58.65±1.19 & 59.40±1.79 & \bfseries 59.99±2.19 \\
\bottomrule
\end{tabular}

  \vspace*{1em}

  \begin{tabular}{llll}
\toprule
{Sexuality} & {Baseline} & {Soc-Dem.} & {Random} \\
\midrule
Bisexual & \bfseries 71.83±1.14 & 71.42±1.51 & 69.46±1.95 \\
Heterosexual & 63.25±0.39 & 63.32±1.21 & \bfseries 63.82±0.55 \\
Homosexual & 64.43±1.75 & \bfseries 66.11±2.20 & 65.12±1.94 \\
\bottomrule
\end{tabular}

  \caption{Average and standard deviation of macro $F_{1}$ from three runs of four-fold stratified cross-validation. Separate table for each attribute. Bold results are the highest averages per group. However, no difference is statistically significant (see Appendix \ref{appendix:significance_tests})}
  \label{tab:results}
\end{table}

\section{Discussion}
\label{sec:discussion}

We do not find strong evidence that explicitly modelling sociodemographics helps to predict annotation behaviour with multi-annotator models.
These results might seem counter-intuitive, given the evidence of systematic annotation differences between sociodemographic groups (see \S\ref{sociodemographics_in_annotation}). This discrepancy, however, echoes the issue highlighted by ecological fallacies \citep{robinson_ecological_1950}:  Not every annotator will be a perfect representative of their group, so we will not necessarily learn additional information based on their group identity. This seems especially true if we already have access to individual behaviour (i.e., individual annotations). 

In contrast to \citet{davani-etal-2022-dealing}, we made sociodemographic information explicit in our experiments, as one of the factors influencing annotation behaviour. Group-specific layers can be seen as an inductive bias putting emphasis on the sociodemographic relations between annotators. However, there are potentially many other factors influencing annotation behaviour (e.g., attitudes, moral values, cognitive biases, psychological traits). In light of our results, it seems plausible that multi-annotator models learn about these factors implicitly as part of predicting individual behaviour, so that making one factor explicit does not change prediction quality, at least in the case of sociodemographics.

Still, we also know that generally group attributes can help predict individual decisions, i.e., as base rates or priors. To avoid ecological fallacies in modelling annotation, we therefore need to
better understand when and how modelling sociodemographic information is useful in predicting an individual annotator's decisions.
For example, we have only evaluated group-specific layers for single attributes.
In contrast, social scientists have long adopted the idea of intersectionality \citep{crenshaw_demarginalizing_1989}, which also informs research on fairness in machine learning \citep{wang_towards_2022}. Intersectionality means that the effect of interactions between sociodemographic attributes enables specific experiences that are not captured by the attributes in isolation. For example, identifying as a man means something different depending on the person's education. 
Groups derived from single attributes might simply be too coarse to improve classifiers learnt from individual labels, as in multi-annotator models.

The dataset we use \citep{kumar_designing_2021} has many characteristics which are ideal for our study (see \S\ref{sec:data}). However, it uses a broad notion of toxicity, in contrast to other studies of toxic language \citep{larimore-etal-2021-reconsidering,sap-etal-2022-annotators}, which match content and analysed groups. When modeling the groups frequently referenced in the datasets themselves, we would expect greater benefits from group-specific layers. Similar to us, \citet{biester-etal-2022-analyzing} who do not find significant differences between annotators of different genders, do so in a more general setting.

We can only partially compare to \citet{gordon_jury_2022}, despite using the same dataset. In addition to differences in approach (see \S\ref{sec:predicting_individual_annotators}), our and their work also differ in their research questions and thus experimental conditions.
\citet{gordon_jury_2022} compare their full model (group and individual) against using \emph{group} information alone. We compare our full model (group and individual) against using \emph{individual} information alone.
So it is unclear if their model would benefit from group information in comparison to individual-level information alone. While they find an improvement from group information it is only in comparison to a baseline predicting not individual but aggregated labels. Additionally, the composition of test sets sampled from the full dataset differs between the studies: \citet{gordon_jury_2022} use a test set of 5,000 comments, while we use 22,360 comments in a four-fold cross-validation. We leave an explicit comparison to future work.

Group-specific layers (\S\ref{sec:experiments}) are a natural extension of annotator-specific classification layers in multi-annotator models. However, other architectures to predict annotator-level labels use different ways to represent sociodemographic information, e.g., via embeddings in a recommender system \citep{gordon_jury_2022}. Future work could explore additional representations of annotator attributes (e.g., as part of the input, either textual or as separate features) and other approaches to modelling the relation of individual labeling decisions and attributes (e.g., probabilistic graphical models).

\section{Conclusion}

We ask how relevant modelling explicit sociodemographic information is in learning from individual annotators.
Our experiments with group-specific layers for four sociodemographic attributes on social media data with toxicity annotations \citep{kumar_designing_2021} show no significant benefit of modelling sociodemographic groups in multi-annotator models.
However, as the issue of ecological fallacies highlights, it is not implausible that these models do not learn additional information from group information beyond the inherent variation.
However, our results do not refute the usefulness of sociodemographic attributes in modelling annotation, but underscore the importance of their judicious use. Different tasks and model architectures will likely benefit to different extents. Ultimately, annotation behaviour is driven by complex factors and we will need to consider more than annotators' sociodemographics.

\section*{Acknowledgements}
We thank Deepak Kumar for providing access to the disaggregated dataset and his continued support. We also thank Aida Mostafazadeh Davani for providing information on implementation details of multi-annotator models. Members of MilaNLP (Bocconi) and the Semantic Computing Group (Bielefeld) provided feedback on earlier versions of this paper, for which we thank them again.

This work has in part been funded by the European Research Council (ERC) under the European Union's Horizon 2020 research and innovation program (grant agreement No. 949944, INTEGRATOR). Likewise, this work has in part been funded by the VolkswagenStiftung as part of the "3B Bots Building Bridges" project.

\section*{Limitations}

While the dataset by \citet{kumar_designing_2021} enabled us to test models for a range of often overlooked groups (e.g., non-binary or bisexual annotators), we ultimately modelled only four specific attributes (gender, age, education, sexual orientation). There are likely to be more factors that could play a role.
Additionally, annotators in the \citet{kumar_designing_2021} dataset are exclusively from the United States of America, so that results do not necessarily hold for other countries or cultures \cite{hovy-yang-2021-importance}. Specifically perceptions of harmful content online are known to vary across countries \citep{jiang_understanding_2021}.

We used only the \citep{kumar_designing_2021} dataset. This is mainly due to our strict criteria regarding dataset size and availability of annotator-level labels and sociodemographic information. These characteristics were a prerequisite for our experiments across different attributes with sufficient numbers of annotators. Most datasets which include annotator-level labels and sociodemographic information contain much smaller numbers of annotators and attributes. Nevertheless, with the \emph{Measuring Hate Speech Corpus} there is at least one additional dataset \citep{sachdeva-etal-2022-measuring} with comparable characteristics that could be used in future experiments. Also, additional small-scale, more focused experiments could use datasets like \citet{sap-etal-2022-annotators} or \emph{HS-Brexit} \citep{akhtar2021opinions} which was annotated by 6 annotators, each from one of two sociodemographic groups.

We do not study the aggregation of individual predictions or evaluate against majority labels, as these are not directly relevant to our investigation of sociodemographic attributes in models of annotation behaviour. Consequently, we cannot derive a conclusion about performance in those settings from our results. This is a noteworthy limitation, because part of the experiments introducing multi-annotator models in \citet{davani-etal-2022-dealing} compare labels aggregated from multi-annotator models against predictions from a standard classifier (directly trained on aggregated labels).

For computational reasons, our experiments use a comparatively small pre-trained language model (RoBERTa, \citealt{liu2019roberta}). 
Thus, results might differ with larger models.

\section*{Ethics Statement}

As sociodemographic attributes are sensitive information, we do not infer attributes, but build on a self-reported, IRB-reviewed dataset \citep{kumar_designing_2021}. 
We also see potential for a discussion of ``privacy by design'' in modelling human label variation based on our results: There can be circumstances in which knowing more about annotators is not relevant, and indeed might lead to violations of privacy.

As multi-annotator models attempt to capture the preferences of individual annotators, there are valid concerns around privacy and anonymity. As discussed in \citet{davani-etal-2022-dealing}, increasing the annotator count can be one option to reduce privacy risks. We show it is feasible to learn a model for a large number of individual annotators (5002 vs.\ 18 and 82 in their work). But a prerequisite for improved privacy is to apply effective aggregation on top of individual predictions, which we do not study in the present work.

%\section*{Acknowledgements}

\bibliography{anthology,custom}

\begin{thebibliography}{38}
\expandafter\ifx\csname natexlab\endcsname\relax\def\natexlab#1{#1}\fi

\bibitem[{Abercrombie et~al.(2022)Abercrombie, Basile, Tonelli, Rieser, and
  Uma}]{nlperspectives-2022-perspectivist}
Gavin Abercrombie, Valerio Basile, Sara Tonelli, Verena Rieser, and Alexandra
  Uma, editors. 2022.
\newblock \href {https://aclanthology.org/2022.nlperspectives-1.0}
  {\emph{Proceedings of the 1st Workshop on Perspectivist Approaches to NLP
  @LREC2022}}. European Language Resources Association, Marseille, France.

\bibitem[{Akhtar et~al.(2020)Akhtar, Basile, and Patti}]{akhtar_modeling_2020}
Sohail Akhtar, Valerio Basile, and Viviana Patti. 2020.
\newblock \href {https://ojs.aaai.org/index.php/HCOMP/article/view/7473}
  {Modeling annotator perspective and polarized opinions to improve hate speech
  detection}.
\newblock In \emph{Proceedings of the {AAAI} Conference on Human Computation
  and Crowdsourcing}, volume~8, pages 151--154.

\bibitem[{Akhtar et~al.(2021)Akhtar, Basile, and Patti}]{akhtar2021opinions}
Sohail Akhtar, Valerio Basile, and Viviana Patti. 2021.
\newblock \href {http://arxiv.org/abs/2106.15896} {Whose opinions matter?
  perspective-aware models to identify opinions of hate speech victims in
  abusive language detection}.
\newblock Preprint arXiv:2106.15896.

\bibitem[{Al~Kuwatly et~al.(2020)Al~Kuwatly, Wich, and
  Groh}]{al-kuwatly-etal-2020-identifying}
Hala Al~Kuwatly, Maximilian Wich, and Georg Groh. 2020.
\newblock \href {https://doi.org/10.18653/v1/2020.alw-1.21} {Identifying and
  measuring annotator bias based on annotators{'} demographic characteristics}.
\newblock In \emph{Proceedings of the Fourth Workshop on Online Abuse and
  Harms}, pages 184--190, Online. Association for Computational Linguistics.

\bibitem[{Basile et~al.(2021)Basile, Fell, Fornaciari, Hovy, Paun, Plank,
  Poesio, and Uma}]{basile-etal-2021-need}
Valerio Basile, Michael Fell, Tommaso Fornaciari, Dirk Hovy, Silviu Paun,
  Barbara Plank, Massimo Poesio, and Alexandra Uma. 2021.
\newblock \href {https://doi.org/10.18653/v1/2021.bppf-1.3} {We need to
  consider disagreement in evaluation}.
\newblock In \emph{Proceedings of the 1st Workshop on Benchmarking: Past,
  Present and Future}, pages 15--21, Online. Association for Computational
  Linguistics.

\bibitem[{Biester et~al.(2022)Biester, Sharma, Kazemi, Deng, Wilson, and
  Mihalcea}]{biester-etal-2022-analyzing}
Laura Biester, Vanita Sharma, Ashkan Kazemi, Naihao Deng, Steven Wilson, and
  Rada Mihalcea. 2022.
\newblock \href {https://aclanthology.org/2022.nlperspectives-1.2} {Analyzing
  the effects of annotator gender across {NLP} tasks}.
\newblock In \emph{Proceedings of the 1st Workshop on Perspectivist Approaches
  to NLP @LREC2022}, pages 10--19, Marseille, France. European Language
  Resources Association.

\bibitem[{Binns et~al.(2017)Binns, Veale, Van~Kleek, and
  Shadbolt}]{binns_like_2017}
Reuben Binns, Michael Veale, Max Van~Kleek, and Nigel Shadbolt. 2017.
\newblock \href {https://doi.org/10.1007/978-3-319-67256-4_32} {Like trainer,
  like bot? inheritance of bias in algorithmic content moderation}.
\newblock In \emph{Social Informatics}, Lecture Notes in Computer Science,
  pages 405--415. Springer International Publishing.

\bibitem[{Cercas~Curry et~al.(2021)Cercas~Curry, Abercrombie, and
  Rieser}]{cercas-curry-etal-2021-convabuse}
Amanda Cercas~Curry, Gavin Abercrombie, and Verena Rieser. 2021.
\newblock \href {https://doi.org/10.18653/v1/2021.emnlp-main.587}
  {{C}onv{A}buse: Data, analysis, and benchmarks for nuanced abuse detection in
  conversational {AI}}.
\newblock In \emph{Proceedings of the 2021 Conference on Empirical Methods in
  Natural Language Processing}, pages 7388--7403, Online and Punta Cana,
  Dominican Republic. Association for Computational Linguistics.

\bibitem[{Crenshaw(1989)}]{crenshaw_demarginalizing_1989}
Kimberle Crenshaw. 1989.
\newblock \href {https://chicagounbound.uchicago.edu/uclf/vol1989/iss1/8}
  {Demarginalizing the intersection of race and sex: A black feminist critique
  of antidiscrimination doctrine, feminist theory and antiracist politics}.
\newblock \emph{University of Chicago Legal Forum}, 1989(1):Article 8.

\bibitem[{Davani et~al.(2022)Davani, D{\'\i}az, and
  Prabhakaran}]{davani-etal-2022-dealing}
Aida~Mostafazadeh Davani, Mark D{\'\i}az, and Vinodkumar Prabhakaran. 2022.
\newblock \href {https://doi.org/10.1162/tacl_a_00449} {Dealing with
  disagreements: Looking beyond the majority vote in subjective annotations}.
\newblock \emph{Transactions of the Association for Computational Linguistics},
  10:92--110.

\bibitem[{Dror et~al.(2017)Dror, Baumer, Bogomolov, and
  Reichart}]{dror-etal-2017-replicability}
Rotem Dror, Gili Baumer, Marina Bogomolov, and Roi Reichart. 2017.
\newblock \href {https://doi.org/10.1162/tacl_a_00074} {Replicability analysis
  for natural language processing: Testing significance with multiple
  datasets}.
\newblock \emph{Transactions of the Association for Computational Linguistics},
  5:471--486.

\bibitem[{Dror et~al.(2018)Dror, Baumer, Shlomov, and
  Reichart}]{dror-etal-2018-hitchhikers}
Rotem Dror, Gili Baumer, Segev Shlomov, and Roi Reichart. 2018.
\newblock \href {https://doi.org/10.18653/v1/P18-1128} {The hitchhiker{'}s
  guide to testing statistical significance in natural language processing}.
\newblock In \emph{Proceedings of the 56th Annual Meeting of the Association
  for Computational Linguistics (Volume 1: Long Papers)}, pages 1383--1392,
  Melbourne, Australia. Association for Computational Linguistics.

\bibitem[{Excell and Al~Moubayed(2021)}]{excell-al-moubayed-2021-towards}
Elizabeth Excell and Noura Al~Moubayed. 2021.
\newblock \href {https://doi.org/10.18653/v1/2021.gebnlp-1.7} {Towards equal
  gender representation in the annotations of toxic language detection}.
\newblock In \emph{Proceedings of the 3rd Workshop on Gender Bias in Natural
  Language Processing}, pages 55--65, Online. Association for Computational
  Linguistics.

\bibitem[{Fornaciari et~al.(2021)Fornaciari, Uma, Paun, Plank, Hovy, and
  Poesio}]{fornaciari-etal-2021-beyond}
Tommaso Fornaciari, Alexandra Uma, Silviu Paun, Barbara Plank, Dirk Hovy, and
  Massimo Poesio. 2021.
\newblock \href {https://doi.org/10.18653/v1/2021.naacl-main.204} {Beyond black
  {\&} white: Leveraging annotator disagreement via soft-label multi-task
  learning}.
\newblock In \emph{Proceedings of the 2021 Conference of the North American
  Chapter of the Association for Computational Linguistics: Human Language
  Technologies}, pages 2591--2597, Online. Association for Computational
  Linguistics.

\bibitem[{Fornaciari et~al.(2022)Fornaciari, Uma, Poesio, and
  Hovy}]{fornaciari-etal-2022-hard}
Tommaso Fornaciari, Alexandra Uma, Massimo Poesio, and Dirk Hovy. 2022.
\newblock \href {https://doi.org/10.18653/v1/2022.acl-demo.12} {Hard and soft
  evaluation of {NLP} models with {BOO}t{ST}rap {SA}mpling - {B}oo{S}t{S}a}.
\newblock In \emph{Proceedings of the 60th Annual Meeting of the Association
  for Computational Linguistics: System Demonstrations}, pages 127--134,
  Dublin, Ireland. Association for Computational Linguistics.

\bibitem[{Freedman(2015)}]{freedman_ecological_2015}
David~A. Freedman. 2015.
\newblock \href {https://doi.org/10.1016/B978-0-08-097086-8.42117-3}
  {Ecological inference}.
\newblock In James~D. Wright, editor, \emph{International Encyclopedia of the
  Social \& Behavioral Sciences (Second Edition)}, pages 868--870. Elsevier.

\bibitem[{Gordon et~al.(2022)Gordon, Lam, Park, Patel, Hancock, Hashimoto, and
  Bernstein}]{gordon_jury_2022}
Mitchell~L. Gordon, Michelle~S. Lam, Joon~Sung Park, Kayur Patel, Jeff Hancock,
  Tatsunori Hashimoto, and Michael~S. Bernstein. 2022.
\newblock \href {https://doi.org/10.1145/3491102.3502004} {Jury learning:
  Integrating dissenting voices into machine learning models}.
\newblock In \emph{Proceedings of the 2022 {CHI} Conference on Human Factors in
  Computing Systems}, {CHI} '22, pages 1--19. Association for Computing
  Machinery.

\bibitem[{Goyal et~al.(2022)Goyal, Kivlichan, Rosen, and
  Vasserman}]{goyal_is_2022}
Nitesh Goyal, Ian~D. Kivlichan, Rachel Rosen, and Lucy Vasserman. 2022.
\newblock \href {https://doi.org/10.1145/3555088} {Is your toxicity my
  toxicity? exploring the impact of rater identity on toxicity annotation}.
\newblock \emph{Proceedings of the {ACM} on Human-Computer Interaction},
  6:1--28.

\bibitem[{Hovy and Yang(2021)}]{hovy-yang-2021-importance}
Dirk Hovy and Diyi Yang. 2021.
\newblock \href {https://doi.org/10.18653/v1/2021.naacl-main.49} {The
  importance of modeling social factors of language: Theory and practice}.
\newblock In \emph{Proceedings of the 2021 Conference of the North American
  Chapter of the Association for Computational Linguistics: Human Language
  Technologies}, pages 588--602, Online. Association for Computational
  Linguistics.

\bibitem[{Jamison and Gurevych(2015)}]{jamison-gurevych-2015-noise}
Emily Jamison and Iryna Gurevych. 2015.
\newblock \href {https://doi.org/10.18653/v1/D15-1035} {Noise or additional
  information? leveraging crowdsource annotation item agreement for natural
  language tasks.}
\newblock In \emph{Proceedings of the 2015 Conference on Empirical Methods in
  Natural Language Processing}, pages 291--297, Lisbon, Portugal. Association
  for Computational Linguistics.

\bibitem[{Jiang et~al.(2021)Jiang, Scheuerman, Fiesler, and
  Brubaker}]{jiang_understanding_2021}
Jialun~Aaron Jiang, Morgan~Klaus Scheuerman, Casey Fiesler, and Jed~R.
  Brubaker. 2021.
\newblock \href {https://doi.org/10.1371/journal.pone.0256762} {Understanding
  international perceptions of the severity of harmful content online}.
\newblock \emph{{PLOS} {ONE}}, 16(8).

\bibitem[{Kumar et~al.(2021)Kumar, Kelley, Consolvo, Mason, Bursztein,
  Durumeric, Thomas, and Bailey}]{kumar_designing_2021}
Deepak Kumar, Patrick~Gage Kelley, Sunny Consolvo, Joshua Mason, Elie
  Bursztein, Zakir Durumeric, Kurt Thomas, and Michael Bailey. 2021.
\newblock \href
  {https://www.usenix.org/conference/soups2021/presentation/kumar} {Designing
  toxic content classification for a diversity of perspectives}.
\newblock In \emph{Seventeenth Symposium on Usable Privacy and Security
  ({SOUPS} 2021)}, pages 299--318. {USENIX} Association.

\bibitem[{Larimore et~al.(2021)Larimore, Kennedy, Haskett, and
  Arseniev-Koehler}]{larimore-etal-2021-reconsidering}
Savannah Larimore, Ian Kennedy, Breon Haskett, and Alina Arseniev-Koehler.
  2021.
\newblock \href {https://doi.org/10.18653/v1/2021.socialnlp-1.7} {Reconsidering
  annotator disagreement about racist language: Noise or signal?}
\newblock In \emph{Proceedings of the Ninth International Workshop on Natural
  Language Processing for Social Media}, pages 81--90, Online. Association for
  Computational Linguistics.

\bibitem[{Liu et~al.(2019)Liu, Ott, Goyal, Du, Joshi, Chen, Levy, Lewis,
  Zettlemoyer, and Stoyanov}]{liu2019roberta}
Yinhan Liu, Myle Ott, Naman Goyal, Jingfei Du, Mandar Joshi, Danqi Chen, Omer
  Levy, Mike Lewis, Luke Zettlemoyer, and Veselin Stoyanov. 2019.
\newblock \href {http://arxiv.org/abs/1907.11692} {Roberta: A robustly
  optimized bert pretraining approach}.
\newblock Preprint arXiv:1907.11692.

\bibitem[{Pedregosa et~al.(2011)Pedregosa, Varoquaux, Gramfort, Michel,
  Thirion, Grisel, Blondel, Prettenhofer, Weiss, Dubourg, Vanderplas, Passos,
  Cournapeau, Brucher, Perrot, and Duchesnay}]{scikit-learn}
F.~Pedregosa, G.~Varoquaux, A.~Gramfort, V.~Michel, B.~Thirion, O.~Grisel,
  M.~Blondel, P.~Prettenhofer, R.~Weiss, V.~Dubourg, J.~Vanderplas, A.~Passos,
  D.~Cournapeau, M.~Brucher, M.~Perrot, and E.~Duchesnay. 2011.
\newblock Scikit-learn: Machine learning in {P}ython.
\newblock \emph{Journal of Machine Learning Research}, 12:2825--2830.

\bibitem[{Plank(2022)}]{plank-2022-problem}
Barbara Plank. 2022.
\newblock \href {https://aclanthology.org/2022.emnlp-main.731} {The
  {``}problem{''} of human label variation: On ground truth in data, modeling
  and evaluation}.
\newblock In \emph{Proceedings of the 2022 Conference on Empirical Methods in
  Natural Language Processing}, pages 10671--10682, Abu Dhabi, United Arab
  Emirates. Association for Computational Linguistics.

\bibitem[{Plank et~al.(2014)Plank, Hovy, and
  S{\o}gaard}]{plank-etal-2014-learning}
Barbara Plank, Dirk Hovy, and Anders S{\o}gaard. 2014.
\newblock \href {https://doi.org/10.3115/v1/E14-1078} {Learning part-of-speech
  taggers with inter-annotator agreement loss}.
\newblock In \emph{Proceedings of the 14th Conference of the {E}uropean Chapter
  of the Association for Computational Linguistics}, pages 742--751,
  Gothenburg, Sweden. Association for Computational Linguistics.

\bibitem[{Prabhakaran et~al.(2021)Prabhakaran, Mostafazadeh~Davani, and
  Diaz}]{prabhakaran-etal-2021-releasing}
Vinodkumar Prabhakaran, Aida Mostafazadeh~Davani, and Mark Diaz. 2021.
\newblock \href {https://doi.org/10.18653/v1/2021.law-1.14} {On releasing
  annotator-level labels and information in datasets}.
\newblock In \emph{Proceedings of the Joint 15th Linguistic Annotation Workshop
  (LAW) and 3rd Designing Meaning Representations (DMR) Workshop}, pages
  133--138, Punta Cana, Dominican Republic. Association for Computational
  Linguistics.

\bibitem[{Robinson(1950)}]{robinson_ecological_1950}
W.~S. Robinson. 1950.
\newblock \href {https://doi.org/10.2307/2087176} {Ecological correlations and
  the behavior of individuals}.
\newblock \emph{American Sociological Review}, 15(3):351--357.

\bibitem[{Röttger et~al.(2022)Röttger, Vidgen, Hovy, and
  Pierrehumbert}]{roettger-etal-2022-two}
Paul Röttger, Bertie Vidgen, Dirk Hovy, and Janet Pierrehumbert. 2022.
\newblock \href {https://doi.org/10.18653/v1/2022.naacl-main.13} {Two
  contrasting data annotation paradigms for subjective {NLP} tasks}.
\newblock In \emph{Proceedings of the 2022 Conference of the North American
  Chapter of the Association for Computational Linguistics: Human Language
  Technologies}, pages 175--190, Seattle, United States. Association for
  Computational Linguistics.

\bibitem[{Sachdeva et~al.(2022)Sachdeva, Barreto, Bacon, Sahn, von Vacano, and
  Kennedy}]{sachdeva-etal-2022-measuring}
Pratik Sachdeva, Renata Barreto, Geoff Bacon, Alexander Sahn, Claudia von
  Vacano, and Chris Kennedy. 2022.
\newblock \href {https://aclanthology.org/2022.nlperspectives-1.11} {The
  measuring hate speech corpus: Leveraging rasch measurement theory for data
  perspectivism}.
\newblock In \emph{Proceedings of the 1st Workshop on Perspectivist Approaches
  to NLP @LREC2022}, pages 83--94, Marseille, France. European Language
  Resources Association.

\bibitem[{Sanh et~al.(2019)Sanh, Debut, Chaumond, and
  Wolf}]{sanh_distilbert_2019}
Victor Sanh, Lysandre Debut, Julien Chaumond, and Thomas Wolf. 2019.
\newblock {DistilBERT, a distilled version of BERT: smaller, faster, cheaper
  and lighter}.
\newblock In \emph{5th Workshop on Energy Efficient Machine Learning and
  Cognitive Computing @ NeurIPS 2019}.

\bibitem[{Sap et~al.(2019)Sap, Card, Gabriel, Choi, and
  Smith}]{sap-etal-2019-risk}
Maarten Sap, Dallas Card, Saadia Gabriel, Yejin Choi, and Noah~A. Smith. 2019.
\newblock \href {https://doi.org/10.18653/v1/P19-1163} {The risk of racial bias
  in hate speech detection}.
\newblock In \emph{Proceedings of the 57th Annual Meeting of the Association
  for Computational Linguistics}, pages 1668--1678, Florence, Italy.
  Association for Computational Linguistics.

\bibitem[{Sap et~al.(2022)Sap, Swayamdipta, Vianna, Zhou, Choi, and
  Smith}]{sap-etal-2022-annotators}
Maarten Sap, Swabha Swayamdipta, Laura Vianna, Xuhui Zhou, Yejin Choi, and
  Noah~A. Smith. 2022.
\newblock \href {https://doi.org/10.18653/v1/2022.naacl-main.431} {Annotators
  with attitudes: How annotator beliefs and identities bias toxic language
  detection}.
\newblock In \emph{Proceedings of the 2022 Conference of the North American
  Chapter of the Association for Computational Linguistics: Human Language
  Technologies}, pages 5884--5906, Seattle, United States. Association for
  Computational Linguistics.

\bibitem[{Shen and Rose(2021)}]{shen-rose-2021-sounds}
Qinlan Shen and Carolyn Rose. 2021.
\newblock \href {https://doi.org/10.18653/v1/2021.eacl-main.152} {What sounds
  {``}right{''} to me? experiential factors in the perception of political
  ideology}.
\newblock In \emph{Proceedings of the 16th Conference of the European Chapter
  of the Association for Computational Linguistics: Main Volume}, pages
  1762--1771, Online. Association for Computational Linguistics.

\bibitem[{Uma et~al.(2021)Uma, Fornaciari, Hovy, Paun, Plank, and
  Poesio}]{uma_learning_survey_2021}
Alexandra~N. Uma, Tommaso Fornaciari, Dirk Hovy, Silviu Paun, Barbara Plank,
  and Massimo Poesio. 2021.
\newblock \href {https://doi.org/10.1613/jair.1.12752} {Learning from
  disagreement: A survey}.
\newblock \emph{Journal of Artificial Intelligence Research}, 72:1385--1470.

\bibitem[{Wang et~al.(2022)Wang, Ramaswamy, and
  Russakovsky}]{wang_towards_2022}
Angelina Wang, Vikram~V Ramaswamy, and Olga Russakovsky. 2022.
\newblock \href {https://doi.org/10.1145/3531146.3533101} {Towards
  intersectionality in machine learning: Including more identities, handling
  underrepresentation, and performing evaluation}.
\newblock In \emph{2022 {ACM} Conference on Fairness, Accountability, and
  Transparency}, {FAccT} '22, pages 336--349. Association for Computing
  Machinery.

\bibitem[{Wolf et~al.(2020)Wolf, Debut, Sanh, Chaumond, Delangue, Moi, Cistac,
  Rault, Louf, Funtowicz, Davison, Shleifer, von Platen, Ma, Jernite, Plu, Xu,
  Le~Scao, Gugger, Drame, Lhoest, and Rush}]{wolf-etal-2020-transformers}
Thomas Wolf, Lysandre Debut, Victor Sanh, Julien Chaumond, Clement Delangue,
  Anthony Moi, Pierric Cistac, Tim Rault, Remi Louf, Morgan Funtowicz, Joe
  Davison, Sam Shleifer, Patrick von Platen, Clara Ma, Yacine Jernite, Julien
  Plu, Canwen Xu, Teven Le~Scao, Sylvain Gugger, Mariama Drame, Quentin Lhoest,
  and Alexander Rush. 2020.
\newblock \href {https://doi.org/10.18653/v1/2020.emnlp-demos.6} {Transformers:
  State-of-the-art natural language processing}.
\newblock In \emph{Proceedings of the 2020 Conference on Empirical Methods in
  Natural Language Processing: System Demonstrations}, pages 38--45, Online.
  Association for Computational Linguistics.

\end{thebibliography}
\bibliographystyle{acl_natbib}

\appendix

\section{Appendix}
\subsection{Annotator Sociodemographics in Sample}
\label{sample-sociodemographics}

Table \ref{tab:annotators} shows how many annotators the sample contains. Counts are given per group of the four attributes gender, age, education and sexuality.

In the \citet{kumar_designing_2021} dataset, sociodemographic attributes are given for each individual annotation - not once per annotator. 
For some annotators, conflicting attribute values exist (e.g., two different age groups). As the data collection spanned several months \citep{kumar_designing_2021}, these value changes can in principle be reasonable (e.g., because an annotator got older, finished a degree, changed sexual preference or gender identity). However, as reasonable changes can not easily be discerned from erroneous input, we disambiguate values based on a heuristic:
If an annotator reports several values for an attribute, we assume the most frequent value to be valid. In cases of no clear most frequent value, we set the attribute to "Prefer not to say". Thus, the main results do not contain annotators with ambiguous attributes.

\begin{table}[h]
  \centering
  \small
  \begin{tabular}{lr}
\toprule
{} &  Number of Annotators \\
Gender            &                       \\
\midrule
Female            &                  2450 \\
Male              &                  2116 \\
Prefer not to say &                   412 \\
Nonbinary         &                    23 \\
Other             &                     1 \\
\bottomrule
\end{tabular}

  \vspace*{1em}
  
  \begin{tabular}{lr}
\toprule
{} &  Number of Annotators \\
Age               &                       \\
\midrule
18 - 24           &                   489 \\
25 - 34           &                  1861 \\
35 - 44           &                  1115 \\
45 - 54           &                   529 \\
55 - 64           &                   321 \\
65 or older       &                   119 \\
Prefer not to say &                   568 \\
\bottomrule
\end{tabular}

  \vspace*{1em}

  \begin{tabular}{lr}
\toprule
{} &  Number of Annotators \\
Sexuality         &                       \\
\midrule
Heterosexual      &                  4018 \\
Bisexual          &                   469 \\
Prefer not to say &                   346 \\
Homosexual        &                   134 \\
Other             &                    35 \\
\bottomrule
\end{tabular}

  \vspace*{1em}

  \begin{tabular}{lr}
\toprule
{} &  Number of Annotators \\
Education           &                       \\
\midrule
Bachelor's degree   &                  1879 \\
College, no degree  &                   861 \\
Prefer not to say   &                   647 \\
Master's degree     &                   642 \\
Associate degree    &                   460 \\
High school         &                   363 \\
Professional degree &                    68 \\
Doctoral degree     &                    51 \\
Below high school   &                    25 \\
Other               &                     6 \\
\bottomrule
\end{tabular}

  \caption{Number of annotators per group for attributes gender, age, sexuality and education. Counts refer to the entire sample}
  \label{tab:annotators}
\end{table}

\subsection{Significance Tests}
\label{appendix:significance_tests}

Results of a replicability analysis \citep{dror-etal-2017-replicability} testing for significant differences in macro $F_{1}$ on scores from three runs of four-fold cross-validation. Table \ref{tab:replication_baseline} shows results for a comparison of the sociodemographic models against the \emph{baseline} models. Table \ref{tab:replication_randomized} shows results for a comparison of the sociodemographic models against the \emph{randomized assignment} models. The Bonferroni correction for the corrected count of significant folds $\hat{k}_{Bonferroni}$ is used to account for the fact that we have overlapping test sets from multiple runs of four-fold cross-validation.

\begin{table}[h]
  \centering
  \small
  \begin{tabular}{lrr}
\toprule
{} &  $\hat{k}_{count}$ &  $\hat{k}_{Bonf.}$ \\
\midrule
Female    &                  2 &                  0 \\
Male      &                  0 &                  0 \\
Nonbinary &                  1 &                  0 \\
\bottomrule
\end{tabular}

  \vspace*{1em}
  
  \begin{tabular}{lrr}
\toprule
{} &  $\hat{k}_{count}$ &  $\hat{k}_{Bonf.}$ \\
\midrule
18 - 24     &                  2 &                  0 \\
25 - 34     &                  2 &                  0 \\
35 - 44     &                  1 &                  0 \\
45 - 54     &                  0 &                  0 \\
55 - 64     &                  1 &                  0 \\
65 or older &                  1 &                  0 \\
\bottomrule
\end{tabular}

  \vspace*{1em}
  
  \begin{tabular}{lrr}
\toprule
{} &  $\hat{k}_{count}$ &  $\hat{k}_{Bonf.}$ \\
\midrule
Bisexual     &                  2 &                  0 \\
Heterosexual &                  4 &                  2 \\
Homosexual   &                  1 &                  0 \\
\bottomrule
\end{tabular}

  \vspace*{1em}
  
  \begin{tabular}{lrr}
\toprule
{} &  $\hat{k}_{count}$ &  $\hat{k}_{Bonf.}$ \\
\midrule
Associate degree    &                  0 &                  0 \\
Bachelor's degree   &                  1 &                  0 \\
Doctoral degree     &                  2 &                  0 \\
High school         &                  0 &                  0 \\
Belowhigh school    &                  0 &                  0 \\
Master's degree     &                  0 &                  0 \\
Professional degree &                  0 &                  0 \\
College, no degree  &                  2 &                  2 \\
\bottomrule
\end{tabular}

  \caption{Results of a replicability analysis of \textit{baseline} vs sociodemographic models. Raw and Bonferroni-corrected counts of significant folds out of 12 folds from three runs of four-fold cross-validation. P-values for each fold are computed via a paired boostrap test with significance level $\alpha = 0.05$, 1000 bootstrap samples per fold and a sample size of 50\% of the respective test set.}
  \label{tab:replication_baseline}
\end{table}

\begin{table}[h]
  \centering
  \small
  \begin{tabular}{lrr}
\toprule
{} &  $\hat{k}_{count}$ &  $\hat{k}_{Bonf.}$ \\
\midrule
Female    &                  2 &                  2 \\
Male      &                  1 &                  0 \\
Nonbinary &                  1 &                  0 \\
\bottomrule
\end{tabular}

  \vspace*{1em}
  
  \begin{tabular}{lrr}
\toprule
{} &  $\hat{k}_{count}$ &  $\hat{k}_{Bonf.}$ \\
\midrule
18 - 24     &                  1 &                  0 \\
25 - 34     &                  0 &                  0 \\
35 - 44     &                  1 &                  0 \\
45 - 54     &                  1 &                  0 \\
55 - 64     &                  3 &                  0 \\
65 or older &                  1 &                  0 \\
\bottomrule
\end{tabular}

  \vspace*{1em}
  
  \begin{tabular}{lrr}
\toprule
{} &  $\hat{k}_{count}$ &  $\hat{k}_{Bonf.}$ \\
\midrule
Bisexual     &                  6 &                  2 \\
Heterosexual &                  1 &                  1 \\
Homosexual   &                  0 &                  0 \\
\bottomrule
\end{tabular}

  \vspace*{1em}

  \begin{tabular}{lrr}
\toprule
{} &  $\hat{k}_{count}$ &  $\hat{k}_{Bonf.}$ \\
\midrule
Associate degree    &                  2 &                  0 \\
Bachelor's degree   &                  1 &                  0 \\
Doctoral degree     &                  0 &                  0 \\
High school         &                  2 &                  0 \\
Belowhigh school    &                  2 &                  0 \\
Master's degree     &                  0 &                  0 \\
Professional degree &                  0 &                  0 \\
College, no degree  &                  1 &                  1 \\
\bottomrule
\end{tabular}

  \caption{Results from replicability analysis of \textit{randomized} vs sociodemographic models. Raw and Bonferroni-corrected counts of significant folds out of 12 folds from three runs of four-fold cross-validation. P-values for each fold are computed via a paired boostrap test with significance level $\alpha = 0.05$, 1000 bootstrap samples per fold and a sample size of 50\% of the respective test set.}
  \label{tab:replication_randomized}
\end{table}

\subsection{Training Details, Hyperparameters and Computational Resources}
\label{appendix:training_details}

We implement models and the training loop using the Hugging Face Transformers library 
(version 4.19.2, \citealt{wolf-etal-2020-transformers}).
Maximum sequence length is 512 tokens, with truncation and padding to the maximum length.
We train for 3 epochs with a batch size of 8 and an initial learning rate of $0.00001$. 
Otherwise, we used default parameters. We found results to particularly depend on the learning rate, 
with higher or lower values leading to worse results. 

We use a weighted loss function. Label weights are calculated per annotator on the training set of each fold. Label weights, evaluation scores and the four-fold dataset splits (StratifiedKFold) are calculated using the scikit-learn library (version 1.0.2, \citealt{scikit-learn}). The folds are based on a fixed random seed per iteration: \texttt{2803636207, 165043843, 2923262358}

The majority of parameters in our model belong to the pre-trained language model shared between all group-specific and annotator-specific layers. Specifically, RoBERTa \citep{liu2019roberta}
in the \emph{roberta-base} variant has 125 Million parameters. We keep the pre-trained model's default output dimensionality of 768, so that each group-specific layer adds $768 * 768 + 768 = 590,592$ parameters and each annotator layer adds $768 * 2 + 2 = 1,538$ parameters.

All experiments ran on a single GPU (GeForce GTX 1080 Ti, 12GB GPU RAM). Per fold, training and evaluation together take about three and a half hours in our setting. Three runs of four-fold cross-validation (12 folds), thus take around 42 hours (1.75 days). With four attributes and three trainable models the combined run time of the reported experiments is estimated to be 21 days. Including preliminary experiments, which, however, mostly were not full runs of k-fold cross-validation and also utilized DistilBERT \citep{sanh_distilbert_2019} with slightly faster run times, it will be many times more. There is no discernible difference in experiment run times between multi-annotator models with or without groups or different numbers of groups.

\subsection{Number of Annotations per Group across all Test Sets}
\label{appendix:support}

Table \ref{tab:support} contains the number of annotations we have per group across the total of 12 folds (from three runs of four-fold cross-validation). This number of annotations is the effective test set size per group. As the numbers do not vary substantially, performance on each fold is equally representative for all groups.

\begin{table}[h]
  \centering
  \small
  \begin{tabular}{llrr}
\toprule
{} & Number Of Annotations &      Min &      Max \\
Gender            &                       &          &          \\
\midrule
Female            &           13555±86.44 &  13383.0 &  13664.0 \\
Male              &           11925±61.65 &  11843.0 &  12062.0 \\
Nonbinary         &              115±6.03 &    104.0 &    122.0 \\
Other             &                5±1.95 &      2.0 &      8.0 \\
Prefer not to say &            2345±51.19 &   2281.0 &   2453.0 \\
\bottomrule
\end{tabular}

  \vspace*{1em}
  
  \begin{tabular}{llrr}
\toprule
{} & Number Of Annotations &    Min &    Max \\
Age               &                       &        &        \\
\midrule
18 - 24           &            2615±50.88 &   2521 &   2697 \\
25 - 34           &           10315±61.45 &  10244 &  10457 \\
35 - 44           &            6250±51.06 &   6179 &   6324 \\
45 - 54           &            3025±47.23 &   2929 &   3083 \\
55 - 64           &            1865±25.48 &   1831 &   1903 \\
65 or older       &             675±19.31 &    643 &    704 \\
Prefer not to say &            3200±55.28 &   3131 &   3289 \\
\bottomrule
\end{tabular}

  \vspace*{1em}

  \begin{tabular}{llrr}
\toprule
{} & Number Of Annotations &    Min &    Max \\
Sexuality         &                       &        &        \\
\midrule
Bisexual          &            2445±39.26 &   2383 &   2501 \\
Heterosexual      &           22630±63.00 &  22507 &  22726 \\
Homosexual        &             725±26.57 &    670 &    759 \\
Other             &              190±7.91 &    173 &    201 \\
Prefer not to say &            1955±35.39 &   1878 &   2009 \\
\bottomrule
\end{tabular}

  \vspace*{1em}

  \begin{tabular}{llrr}
\toprule
{} & Number Of Annotations &    Min &    Max \\
Education           &                       &        &        \\
\midrule
Associate degree    &            2605±47.59 &   2516 &   2697 \\
Bachelor's degree   &           10510±84.79 &  10348 &  10700 \\
Doctoral degree     &             305±18.83 &    270 &    332 \\
High school         &            2080±37.01 &   2015 &   2139 \\
Below high school   &             165±11.17 &    144 &    184 \\
Master's degree     &            3515±48.08 &   3425 &   3580 \\
Other               &               30±3.44 &     25 &     36 \\
Prefer not to say   &            3690±52.92 &   3603 &   3808 \\
Professional degree &             380±17.87 &    352 &    411 \\
College, no degree  &            4665±71.36 &   4539 &   4776 \\
\bottomrule
\end{tabular}

  \caption{Average, standard deviation, minimum and maximum of number of annotations per fold. All information given per group of gender, age, education and sexuality. Statistics are calculated across 12 folds from three runs of four-fold cross-validation.}
  \label{tab:support}
\end{table}

\subsection{Full Results}
\label{app:full_results}
Table \ref{tab:full_results} shows full results of experiments (see \ref{sec:experiments}), including results for all residual categories and a naive baseline which always predicts \textit{toxic}. 

\setlength{\tabcolsep}{2pt}
\begin{table*}[ht]
  \centering
  \small
  \begin{tabular}{lllll}
\toprule
{Gender} & {Majority Baseline} & {Baseline} & {Soc-Dem.} & {Random} \\
\midrule
Female & 41.79±0.12 & 62.23±0.53 & 62.25±1.19 & \bfseries 62.41±0.92 \\
Male & 40.53±0.11 & \bfseries 68.00±0.49 & 67.66±0.46 & 67.63±0.53 \\
Nonbinary & 44.69±1.39 & 56.33±6.00 & 56.80±7.24 & \bfseries 58.00±7.49 \\
Other & 45.50±4.69 & 48.56±10.78 & \bfseries 50.53±14.63 & 43.66±7.25 \\
Prefer not to say & 41.05±0.36 & 64.54±1.13 & 65.05±1.52 & \bfseries 65.08±1.86 \\
\bottomrule
\end{tabular}

  \vspace*{1em}

  \begin{tabular}{lllll}
\toprule
{Age} & {Majority Baseline} & {Baseline} & {Soc-Dem.} & {Random} \\
\midrule
18 - 24 & 42.49±0.28 & 59.39±1.58 & 60.44±1.05 & \bfseries 60.52±1.37 \\
25 - 34 & 40.49±0.09 & 66.72±0.56 & 66.63±0.83 & \bfseries 66.92±0.51 \\
35 - 44 & 41.87±0.15 & 64.50±0.59 & 64.94±1.33 & \bfseries 65.24±0.89 \\
45 - 54 & 40.63±0.26 & 65.68±0.66 & 65.88±1.39 & \bfseries 65.98±0.83 \\
55 - 64 & 41.65±0.39 & 64.37±1.22 & \bfseries 64.94±1.66 & 64.84±1.30 \\
65 or older & 41.46±0.54 & 63.34±2.07 & \bfseries 64.70±2.21 & 62.77±2.39 \\
Prefer not to say & 41.37±0.32 & 63.99±1.32 & \bfseries 65.24±1.18 & 64.73±1.33 \\
\bottomrule
\end{tabular}

   \vspace*{1em}

  \begin{tabular}{lllll}
\toprule
{Education} & {Majority Baseline} & {Baseline} & {Soc-Dem.} & {Random} \\
\midrule
Associate degree & 43.16±0.19 & 60.69±1.44 & 60.54±2.35 & \bfseries 60.78±1.62 \\
Bachelor's degree & 40.38±0.10 & 66.16±0.51 & 66.23±0.82 & \bfseries 66.80±0.54 \\
Doctoral degree & 43.34±0.94 & 61.93±3.82 & \bfseries 63.79±5.03 & 63.27±3.67 \\
High school & 43.02±0.26 & 60.53±1.39 & 60.47±2.22 & \bfseries 60.55±1.87 \\
Below high school & 43.10±1.44 & 58.28±4.68 & \bfseries 62.12±4.90 & 60.17±4.25 \\
Master's degree & 37.55±0.32 & \bfseries 69.71±0.86 & 69.58±0.93 & 69.45±0.96 \\
Other & 42.95±2.31 & 56.56±10.88 & 57.59±9.86 & \bfseries 57.71±12.28 \\
Prefer not to say & 40.97±0.27 & 65.07±1.16 & 65.69±1.05 & \bfseries 65.74±1.09 \\
Professional degree & 40.43±0.80 & 66.75±2.37 & 67.84±3.32 & \bfseries 68.62±2.84 \\
College, no degree & 43.61±0.18 & 58.65±1.19 & 59.40±1.79 & \bfseries 59.99±2.19 \\
\bottomrule
\end{tabular}

  \vspace*{1em}

  \begin{tabular}{lllll}
\toprule
{Sexuality} & {Majority Baseline} & {Baseline} & {Soc-Dem.} & {Random} \\
\midrule
Bisexual & 34.69±0.50 & \bfseries 71.83±1.14 & 71.42±1.51 & 69.46±1.95 \\
Heterosexual & 41.99±0.06 & 63.25±0.39 & 63.32±1.21 & \bfseries 63.82±0.55 \\
Homosexual & 41.15±0.41 & 64.43±1.75 & \bfseries 66.11±2.20 & 65.12±1.94 \\
Other & 43.53±0.78 & 57.55±3.79 & \bfseries 60.57±4.51 & 58.69±4.72 \\
Prefer not to say & 39.12±0.24 & \bfseries 67.80±1.56 & 67.27±1.52 & 67.46±1.11 \\
\bottomrule
\end{tabular}

  \caption{Average and standard deviation of macro $F_{1}$ from three runs of four-fold stratified cross-validation. Separate table for each attribute. Bold results are the highest average per group. Full results including naive majority baseline}
  \label{tab:full_results}
\end{table*}

\end{document}